# Leveraging Simultaneous Usage of Edge GPU Hardware Engines for Video Face Detection and Recognition

Asma Baobaid and Mahmoud Meribout, Senior *Member, IEEE*

*Abstract* —Video face detection and recognition in public places at the edge is required in several applications, such as security reinforcement and contactless access to authorized venues. This paper aims to maximize the simultaneous usage of hardware engines available in edge GPUs nowadays by leveraging the concurrency and pipelining of tasks required for face detection and recognition. This also includes the video decoding task, which is required in most face monitoring applications as the video streams are usually carried via Gbps Ethernet network. This constitutes an improvement over previous works where the tasks are usually allocated to a single engine due to the lack of a unified and automated framework that simultaneously explores all hardware engines. In addition, previously, the input faces were usually embedded in still images or within raw video streams that overlook the burst delay caused by the decoding stage. The results on real-life video streams suggest that simultaneously using all the hardware engines available in the recent NVIDIA edge Orin GPU, higher throughput, and a slight saving of power consumption of around 300 mW, accounting for around 5%, have been achieved while satisfying the real-time performance constraint. The performance gets even higher by considering several video streams simultaneously. Further performance improvement could have been obtained if the number of shuffle layers that were created by the tensor RT framework for the face recognition task was lower. Thus, the paper suggests some hardware improvements to the existing edge GPU processors to enhance their performance even higher.

## I. Introduction

Since the late 1980s, video face detection and recognition have attracted the attention of many researchers and companies as they can be used in several applications involving security reinforcement and smart access to authorized zones such as nuclear plants. The associated algorithms include local phase quantization (LPO) [1.], local binary patterns (LBP) [2], or patterns of oriented edge magnitudes (POEM) [3]. They are then processed by higher-level algorithms such as principal component analysis (PCA) [4], support vector machines (SVM) [5], and linear discriminant analysis (LDA) [6]. However, they are relatively weak to be adopted in real-life applications. For instance, they can hardly cope with pose angle variations, rotations, and zooms of faces. The

Asma Baobaid and Mahmoud Meribout are with the Electrical & Computer Engineering Department, Khalifa University, Abu Dhabi, UAE (email: asma.baobaid@outlook.com and mahmoud.meribout@ku.ac.ae)

recent emergence of deep learning algorithms at the edge has contributed to achieving portable smart cameras at the edge that perform high-speed face detection and classification with good accuracy. These algorithms are usually based on Convolutional Neural Networks (CNN), and the associated hardware accelerators are mainly either GPU, TPU, or FPGA-based [5][6]. Nevertheless, most of them still cannot be achieved in real-time, and/or they are not accurate enough for task-critical applications. For instance, state-of-the-art FPGAs do not have enough built-in memory (i.e., Mbits capacity order) and are limited to hosting only light CNN models, which are not accurate enough. Recent Xilinx Versal FPGAs offer more SOC solutions by integrating a TPU, programmable logic (PL), and processing systems (PS) and can be a good candidate for hosting complex CNN models if the built-in memory capacity, which is still in the Mbits range, can be increased further. TPU processors, which are mainly manufactured by Google company, are not flexible enough to support computation models that are not TPU compatible [4]. In addition, they are not commercially available. In this paper, we are interested in edge GPU hardware accelerators that offer more flexibility and large on-chip memory (e.g., in the ranges of Gbits capacity) to host more complex CNN models with reasonable power consumption. Nevertheless, most implementations solely used the GPU core and, to a lesser extent, the CPU core and ignored exploring other built-in power-efficient hardware accelerators. The lack of an automated hardware-software framework and the fact that the scheduling mechanisms are largely undocumented are one of the reasons for this trend [7]. This paper demonstrates that solely using GPU cores causes memory conflict and increases memory cache misses, leading to slightly degrading system performance. The issue becomes even worst for batch processing, where several video streams are processed simultaneously. This paper suggests a parallel hardware implementation of face detection and recognition using simultaneously all the hardware engines available in modern NVIDIA edge GPU. Namely the NVIDIA AGX Orin GPU. Its main contribution can be summarized as follows:

1- Video detection and recognition of faces is implemented in a most recent edge GPU processor to improve the existing systems' latency, throughput, and power consumption. This is an improvement over previous research works, which mainly considered still images or raw video frames and overlooked the delay that may be induced by compulsory hardware modules such as video decoder and multiplexer/demultiplexer and the associated memory access mechanisms. The performance is enhanced by maximizing the usage of different hardware engines available in the recent edge GPU through adequate hardware allocation and scheduling. This achievement can potentially promote the development of other similar emerging video applications, such as real-time machine vision for autonomous vehicles.

2- Leverage batch processing to enhance the system performance even further. Batch processing, which, to the author's knowledge, was not previously tackled for video face detection and recognition on edge GPUs, enhances parallelism by mitigating the memory contention between pipeline stages. The paper demonstrates that handling two video streams simultaneously leads to slightly higher throughput while consuming approximately the same power as a single video stream. The performance could be even slightly better if each video stream were handled by a separate hardware engine, which cannot be done in the current NVIDIA framework.

3- A reasonable number of faces ranging from 3 to 9 faces and taken from public videos [8] in addition to other videos taken in the lab are considered in this paper. This is better than other related works, which have considered only up to 4 faces [9]. In addition, some of the videos used include traditional headscarves from the Middle East region, which were not considered in earlier works.

4- Following the experimental results, the paper highlights some unexpected findings about the used NVIDIA GPU and suggests some hardware improvements for the existing edge GPU processors to enhance their performance even better.

## II. RELATED WORKS

Video face detection and recognition determines the most likely identities existing in an input video stream within at least its current throughput. It consists of two sequential phases: face detection to identify the locations of faces within the current frame, followed by face classification. These two tasks are time-consuming and must be handled by a dedicated hardware accelerator. In the face detection phase, minimum bounding boxes surrounding the faces are generated, leading to a computation time solely proportional to the frame size, irrespective of the number of faces. Each bounding box is scaled and processed separately in the face recognition phase, which makes its computation time complexity proportional to the number of faces in the image. Viola Jones algorithm [10], which is based on Haar feature extraction, is one of the most non-DNN (deep neural network) algorithms used for face detection, while MTCNN (Multi-task cascaded CNN) [11] and NVIDIA's FaceDetect [12] are the most widely used CNN algorithms. These algorithms feature low-level processing tasks, making them suitable for fine-level parallelism architectures featured in FPGAs and GPUs. However, GPU architecture is more adequate as it can accommodate the large number of parameters that need to be stored on-chip for both the MTCNN and FaceDetect models. This is also the case of CNN-based face recognition algorithms such as Google's FaceNet [13]. Several parallel implementations of other non-AI face detection and recognition algorithms on edge devices have already been suggested in the literature. For instance, in [14], the SIFT (scale-invariant feature transform) algorithm was implemented on a cloud GPU (Tesla C2050 with 448 CUDA cores operating at 1.15 GHz) to achieve a low throughput of 11 fps for 1920 x 1440 input video frames. In [15] a face detection implementation using Xilinx Virtex-5 LX330 FPGA is proposed based on LBP, where the implementation achieved a throughout of 307 FPS for 640x480 input size images. Face tracking and recognition using radial basis function neural network (RBFNN) was successfully implemented in [16] on two hardware platforms based on Xinlinx' Zinq FPGA and DSP processors to achieve 14 and 4.5 recognitions per second, respectively. In [17] a multispectral camera operating in the LWIR range was suggested to quickly determine the eyes' centers to proceed with the face recognition. This solution may not work well in a cluttered scene, especially when the faces are away from the camera; it is also very costly and requires special cumbersome arrangements to keep the LWIR camera cool. As was mentioned in [18], compared to legacy machine vision algorithms, deep learning algorithms contributed to significantly improving the accuracy of face detection and recognition while keeping a similar low-level computation model that is well adapted to fine parallelism. Thus, several other hardware accelerators based on CNN algorithms for face detection and recognition were also suggested recently. For instance, in [9], a comparative study of face recognition on both edge and cloud GPU devices revealed that the Jetson AGX Xavier and Jetson NX edge GPUs can achieve face recognition at an average throughput of 20 fps and 12 fps respectively, below the minimum 30 fps. One reason for this underperformance is that the authors did not explore all hardware modules available on the GPU processors, as they exclusively used the CUDA core. In addition, the authors did not explore the pipelining features of the GPUs. This is required as the cameras used for face detection and recognition are usually networked and thus yield encoded video streams. Some researchers suggest quantifying the weights and model pruning to tackle this issue, reducing complexity and memory storage requirements. For instance, in [19], it was found that the 4 and 8-bit versions of MobileFaceNet can lead to an impressive 98.68% and 98.63% accuracy on the CASIA-Webface dataset [20]. However, the authors did not discuss the system throughput and how many faces were tested to achieve such accuracy. Indeed, MobileFaceNet is a tiny CNN model that is not expected to perform well for a large dataset. A similar CNN model, MobileNet, was recently suggested in [21] to be hosted on an FPGA chip for face tracking applications, which is, however, less computation-intensive than face recognition. Very recently, one of the rare works that considered video face detection and recognition suggested using a video centralized transformer. While the method explores the temporal correlation between frames, the authors did not suggest a hardware accelerator as the algorithm is very time-consuming. A possible application sought by the authors is face retrieval in video streams which can accommodate low throughput. A single CMOS face recognition chip that uses a binarized neural network (BNN) was recently suggested to achieve an impressive throughput of 240 fps for 97.75% accuracy and 2.78 mW [22]. However, the shallow structure of the network lets the system consider only single-face images of 120 x 120 pixels. Some systems add depth information to enhance the accuracy using either LIDAR or stereovision [23]. However, this makes the system more cumbersome and costly. Other studies have suggested using simultaneously FPGA, CPU, and GPU to partition and pipeline HAAR algorithm, resulting in lower latency compared to using the GPU code alone, probably due to the memory bottleneck [24]. However, it is unclear how this platform can handle more complex CNN tasks, including the ones used for face detection and recognition. It can be concluded that face detection and recognition hardware accelerators based or not on CNN still lack real-time performance. Indeed, most of the edge GPU-based hardware accelerators for face detection and recognition do not explore all the hardware resources that are available in the GPU. They solely use the CUDA cores and, to a lower extent, the CPU cores (mainly for controlling the CUDA core and managing the memory transfer) and ignore the other valuable hardware engines that can be simultaneously used, such as DLAs (deep learning accelerators), tensor cores, programmable vision accelerator (PVA), and the video encoder/decoder hardware engines. Among the few works that suggested simultaneous usage of GPU resources, one can cite the DeepX framework presented in [25]. The framework allows the splitting of CNN layers into GPU core and CPU cores to reduce the latency of the CNN models. Similarly, in [26], a channel-wise distribution between GPU and CPU was suggested at a fraction that minimizes the algorithm latency. Nevertheless, using DLA, GPU, VPA, and video decoding/encoding engines together in heterogeneous systems for CNN models, including those targeting face detection and recognition applications, has not yet been reported. This paper leverages the parallel use of these hardware engines to implement face detection and recognition tasks.

## III. Parallel GPU-Based Hardware Accelerator

### A. GPU Hardware Platform

Fig. 1(a) shows the hardware block diagram of the Orin GPU processor, the latest edge GPU manufactured by NVIDIA company which was used in this paper [27]. Fig. 1(b) shows the experimental setup used in the paper. The GPU comprises 16 Stream Multiprocessors (SMs) in the CUDA core engine, each comprising 128 single instruction multiple thread (SIMT) units. Also, similarly to any other GPU architectures, the processor features several memory hierarchies, with thread-local registers being the fastest (1 clock latency), followed by the L1 memory cache, which is shared by all SIMT units of the same SM (192 kB/SM), which is enough to store only one video frame fully. Thus, using SIMTs belonging to the same SM to implement pipeline stages may not yield the best possible performance, as this will lead to a high number of memory cache misses. The slowest 4 MB L2 memory cache is shared by all SMs to yield around 200 cycles latency making it possible to store up to two video frames. This can be useful to implement a two pipeline stages architecture within the CUDA core by allocating SIMTs to different SMs. Implementing deeper pipeline architecture can still be effective but would lead to a slight decrease in performance caused by more frequent memory cache misses. The off-chip 256-bit data bus-64 GB LPDRAM5 memory, which yields the longest latency, can store several frames that the SM, DLA, PVA, video encoder, and video decoder engines can access at 204.08 GB/s. This impressive bandwidth can be very useful for multi-batch processing since it far exceeds the bandwidth of a single standard video stream (i.e., 30fps).

Thus, the large number of SIMTs, which are available in the processor allows a high level of parallelism and can, therefore, potentially accelerate tasks that can be split into several independent subtasks, especially when floating point operations are required. This constitutes a great advantage over FPGAs, which can mainly host integer or fixed-point operations if a high level of parallelism is sought. Each SM also includes a tensor core, which performs hardwired matrix multiplications involving half-precision (FP16) or integer operations (INT8). It can operate concurrently with the SIMT of the same SM, as far as there is no memory bottleneck. The DLA hardware is a fixed-function accelerator engine targeted for deep learning operations, which is not available in the recent cloud GPUs such as NVIDIA's Blackwell GPUB200 [26]. It's designed to perform hardware acceleration of convolutional neural networks, supporting various layers such as convolution, deconvolution, fully connected, activation, pooling, and batch normalization.

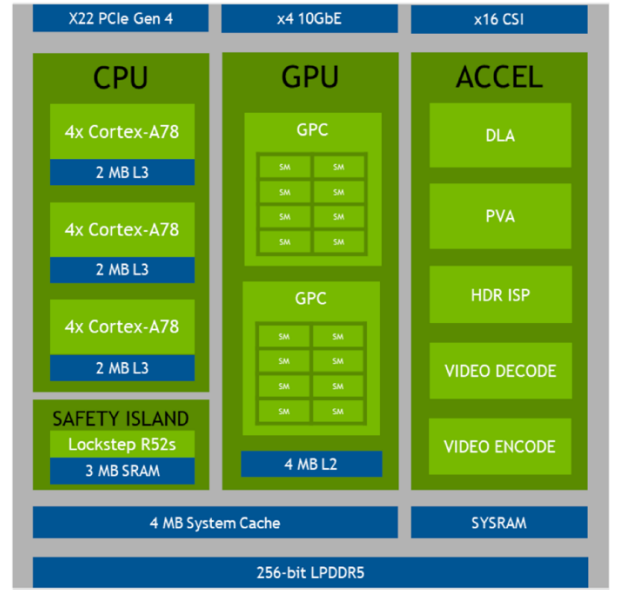

(a)

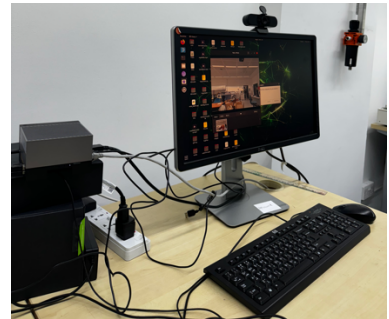

(b)

Fig. 1: Block diagram of the Jetson AGX ORIN GPU (a) and photograph of the setup (b)

Thus, the large number of SIMTs, which are available in the processor allows a high level of parallelism and can, therefore, potentially accelerate tasks that can be split into several independent subtasks, especially when floating point operations are required. This constitutes a great advantage over FPGAs, which can mainly host integer or fixed-point operations if a high level of parallelism is sought. Each SM also includes a tensor core, which performs hardwired matrix multiplications involving half-precision (FP16) or integer operations (INT8). It can operate concurrently with the SIMT of the same SM, as far as there is no memory bottleneck. The DLA hardware is a fixed-function accelerator engine targeted for deep learning operations, which is not available in the recent cloud GPUs such as NVIDIA's Blackwell GPUB200 [26]. It's designed to perform hardware acceleration of convolutional neural networks, supporting various layers such as convolution, deconvolution, fully connected, activation, pooling, and batch normalization. Its main advantage is to yield a performance/W 2.5 times higher than the SMs by computing a total of 2 x 52.5 INT8 sparse TOPs, which is convenient for battery-powered applications. Two DLA engines (version 2.0) and a PVA are available in the ORIN GPU [28]. Despite performing slightly less than the SMs, the DLA engine has the advantage of having a dedicated shared buffer of 2 MB which can typically store fully one video frame and any intermediary layer of a CNN model. This brings a valuable solution to the memory conflict issue when multiple pipeline stages are implemented within the same or different SMs.

### B. Parallel Hardware Algorithm

Fig. 2 shows the suggested hardware algorithm. It consists of 4 pipeline stages that use different hardware engines that are available on edge ORIN GPU hardware. Previous GPU-based face detection and recognition systems did not consider this simultaneous usage of hardware accelerators within the GPU. Currently, most cameras used in public places typically feature an Ethernet interface to generate the output video stream as per some international standards, such as H264, or H265. Thus, performing face detection and recognition at the edge requires a dedicated real-time video decoding engine featured in most recent NVIDIA GPUs, including the ORIN GPU. It can decode up to 6 and 24 H265 video frames in real-time for 4K and 1080 image sizes, respectively. Thus, a single GPU can be interfaced to multiple video cameras through a multiport asymmetric Power on Ethernet (POE) switch with a 10-Mbps bandwidth interface for connecting the cameras (since H265, for instance, yields less than 4 Mbps bit rate) and 1 to 10 Gbps bandwidth for the GPU interface. This constitutes an advantage over FPGA-based hardware accelerators, which usually do not incorporate dedicated video-decoding circuitry. The video decoding stage in Fig. 2, which causes unpredicted but small latency, consumes very low power as it is implemented into a dedicated ASIC chip. The video processing stage preprocesses the decoded video frame by removing eventual noises and resizing the input video frame to match the frame size expected by the face detection CNN model. This is done using the on-chip PVA, which features a VLIW (very long instruction word) architecture to add another latency, which is indeed negligible since the LPDRAM5 memory bandwidth is extremely high. The face detection and recognition algorithms, which are based on FaceDetect [12] and FaceNet [13] models, respectively, are implemented on both the CUDA core and the DLA engine. These models are one of the most accurate algorithms which are available so far. An optional video encoder stage, is shown in Fig. 2, is used to save the output file. However, displaying the output to the screen will not require the encoding stage. This paper uses the same pipeline blocks to host one or multiple video streams using an additional front-end video multiplexer.

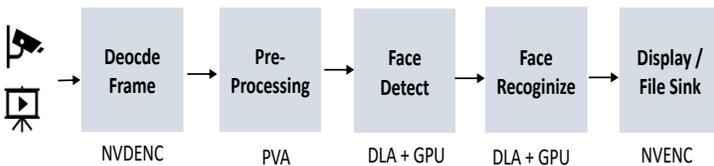

Fig. 2: The suggested pipelined GPU implementation of the face detection/recognition

One hardware allocation strategy considered in this paper was to maximize the DLA utilization and to leave the remaining layers on the CUDA core/tensor core without ignoring the layer's dependencies during the allocation. This would mitigate the memory conflict that may occur between the pipeline stages, which consequently would enhance the overall system throughput compared to using the GPU alone. Fig. 3(b) shows the execution time saving, which can be expected by involving both the DLA and CUDA core in the execution compared to when the model is run sequentially (Fig. 3(a)). The initialization step is allocated for video capturing and decoding using the built-in video decoder engine. The constraint here is that the overall throughput of the algorithm should not exceed one frame acquisition time (i.e., 30 ms). Nevertheless, as of today, DLA alone cannot fully host some CNN models, including the FaceNet model used in this paper. Another allocation strategy is to implement the pipelining using two different SMs or different SIMTs within the same SMs, respectively (Fig. 3(c)). This paper demonstrates that this strategy can lead to a very good performance, however slightly lower than the one offered by the DLA engine, due to the L2 memory cache conflict. The discrepancy in performance gets even higher for the case of multiple video streams as more memory conflict occurs due to the requirement to store multiple video streams into the L2 shared memory that different SMs are required to share. It is worth noting that since the full FaceDetect and FaceNet models are simultaneously implemented on GPU and/or DLA without changing their respective structures, the accuracy is not compromised as in the case of lightweight versions.

### C. Hardware Partitioning Strategies

The FaceDetect model uses ResNet18 for feature extraction (Fig.4(a)). It is based on the NVIDIA DetectNet_v2 detector, which uses bounding-box regressions on a uniform grid in the input image. It initially divides the input frame into a grid that predicts four normalized bounding-box parameters ($x_c$, $y_c$, w, h) and confidence value per output class. The raw normalized bounding box and confidence detections must be processed by a clustering algorithm such as DBSCAN to produce the final bounding-box coordinates and category labels. The model consists of 18 layers, which start with a convolutional layer for an input image of 224 x 224, followed by a pooling layer (maxpool) and 4 convolution blocks, which each contain 4 convolution layers. This is completed by an average pooling layer and a fully connected layer. This structure makes the model suitable for both DLA and GPU implementation. The model architecture of FaceNet is presented in Fig. 4(b). It was originally developed by Google to predict the faces' identities [29]. Similarly to FaceDetect, its backbone network consists of a sequence of convolutional and pooling layers to end with a 128-element vector representation called face-embedding. Face embeddings are then mapped to generate a mapped Euclidean space. One of the main advantages of this model is to require a reasonable 128 bytes/face and only a minimal alignment. However, its disadvantage is that it can only recognize one face at a time, which requires using face detection algorithm in case several faces are expected in an image. In [28], a face recognition system based on FaceNet was demonstrated to achieve an impressive accuracy of 99.63% on the LFW dataset. However, the suitability of the algorithm to run on IoT/edge devices was not investigated and was recommended as a major future goal by the authors.

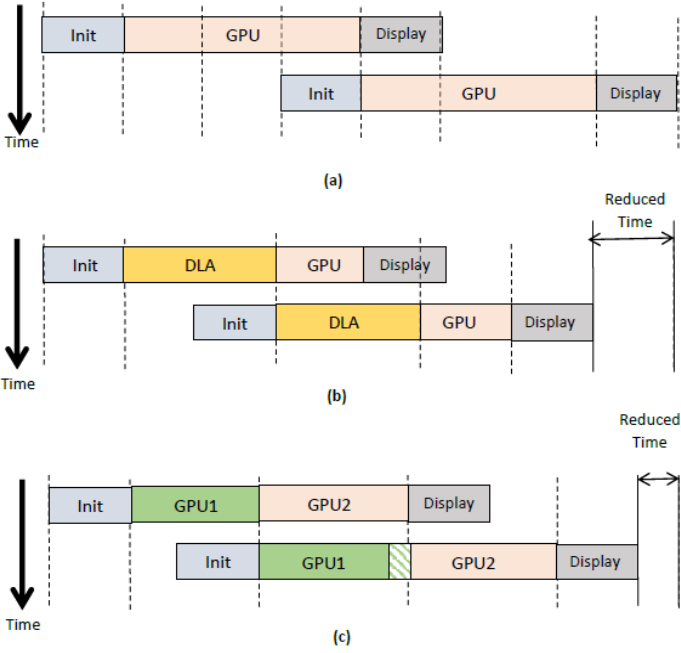

Fig. 3: Example of hardware allocation without using pipelining (a) using the DLA + GPU cores in the pipelining(b) and using SMs only of the GPU core (c)

| Type | Output Size |
|---|---|
| Conv1 (7x7, stride 2) | 112x112x64 |
| Max Pool (3x3, stride 2) | 56x56x64 |
| Residual Block (2x) | 56x56x64 |
| Residual Block (2x) | 28x28x128 |
| Residual Block (2x) | 14x14x256 |
| Residual Block (2x) | 7x7x512 |
| Avg Pool (7x7) | 1x1x512 |
| Fully Connected | 1x1x1000 |
| **Total** | |

(a)

| Type | Output Size |
|---|---|
| conv1 (7x7x3, 2) | 112x112x64 |
| max pool + norm | 56x56x64 |
| inception (2) | 56x56x192 |
| norm + max pool | 28x28x192 |
| inception (3a) | 28x28x256 |
| inception (3b) | 28x28x320 |
| inception (3c) | 14x14x640 |
| inception (4a) | 14x14x640 |
| inception (4b) | 14x14x640 |
| inception (4c) | 14x14x640 |
| inception (4d) | 14x14x640 |
| inception (4e) | 7x7x1024 |
| inception (5a) | 7x7x1024 |
| inception (5b) | 7x7x1024 |
| avg pool | 7x7x1024 |
| fully conn | 1x1x128 |
| L2 normalization | 1x1x128 |
| **Total** | |

(b)

Fig. 4: Structure of FaceDetect (a) and FaceNet (b)

An important design aspect for accelerating the system throughput is the policy to allocate the FaceDetect and FaceNet models onto the GPU and DLA hardware engines. This suggests taking into consideration simultaneously the following constraints:

### C.1 Reduce the hardware engines' ideal time

Minimizing the ideal time of all hardware engines while leveraging concurrency allows for an increase in the system throughput. For instance, if more time is allocated to DLA, then the GPU core is idle most of the time. This suggests determining the computation time of each of the FaceDetect and FaceNet when they are fully implemented onto the GPU and DLA: $C_{GPU}^{FD}$, $C_{DLA}^{FD}$, $C_{GPU}^{FN}$, $C_{DLA}^{FN}$. The Fraction of execution of each model, $\tau_{GPU}^{FD}$, $\tau_{DLA}^{FD}$, $\tau_{GPU}^{FN}$, $\tau_{DLA}^{FN}$ shall

then satisfy the following equations for FaceDetect and FaceNet, respectively:

$$\frac{C_{GPU}^{FD}}{\tau_{GPU}^{FD}} = \frac{C_{DLA}^{FD}}{\tau_{DLA}^{FD}} \tag{1}$$

and

$$\frac{C_{GPU}^{FN}}{\tau_{GPU}^{FN}} = \frac{C_{DLA}^{FN}}{\tau_{DLA}^{FN}} \tag{2}$$

In the case of a single video stream, the use of DLAs and SMs to leverage pipelining and parallelism is proposed as a computation and power-efficient solution. However, as it is mentioned in the literature, without pipelining, when using solely the DLA engine to implement one of the CNN models, the SMs performance is slightly better but at the cost of higher power consumption. Another partitioning solution that paid off in our experiments is to proceed with the hardware allocation at the CNN-model level rather than at the layer level. This can be a good alternative for any two cascade CNN models having approximately the same number of layers, which is the case of FaceDetect and FaceNet. This would reduce the number of memory transfers and avoids using low-level coding required to map each layer into a specific engine. Thus, another room for improvement in current NVIDIA software frameworks is to provide an automated scheduling scheme at the layer level using, for instance, Equations (1) and (2) above. In the case of FaceDetect and FaceNet, L2 memory cache capacity is enough to store a good fraction of the two video frames, which are processed by each of the two SM pipeline stages (e.g. face detection and face recognition pipeline stages) of the same GPU. This lowers the memory cache misses, which has the advantage of enhancing the system throughput.

### C.2 Map supported layers into DLA

In addition to Equations (1) and (2), the hardware allocation of face detection and recognition needs to take into consideration that not all layers of the CNN models can be mapped onto the DLA, while the GPU core engine can accommodate all of them. The unsupported layers by the DLA will have to fall back to the GPU for execution. As was mentioned earlier, The DLA engines support several layers but under some conditions. For instance, despite initially not comprising any shuffle layer, the requirement of using TensorRT to map the Facenet model into DLA causes the creation of several shuffle layers [8]. The purpose of a shuffle layer is to transform the format of the output of a given layer to make it compatible with the DLA hardware architecture. The transforms, which include resizing and transposing, are relatively simple to implement in hardware but are not operated within DLA, forcing them to be mapped back to GPU. This would cause frequent GPU-DLA communications, which alters the system's performance. One solution to overcome this limitation is to provide a shuffle layer accelerator within the DLA engine in next-generation edge GPUs. This allows a seamless implementation of existing CNN models into DLA without compromising the system's performance. On the other hand, FaceDetect does not include any shuffle layer, which allows it to be completely mapped onto the DLA, using, for instance, one of the partitioning schemes shown in Fig. 5(a) and (b) with three (2 sets on DLA and 1 set onto GPU) and five (3 sets on DLA on two sets on GPU) partitions respectively. Increasing the number of partitions has the advantage of yielding deeper pipelining and eventually higher throughout. However, this would also cause more memory transfers between the GPU core and DLA engine (i.e. memcpy task using MemCpyDtoD function). As illustrated in Fig.5 (a), a single layer (layer X) mapped onto the CUDA core would require 2 memory accesses to/from the DLA buffer memory, whereas, in (b), increasing these layers to 2 (layers X and Y) will lead to 4 device-to-device memory transfers. This causes an overhead compared to the SM hardware allocation alone, where the unique memory transfer is between the CPU and the GPU using MemCpyDtoH and MemCpyHtoD functions. The same applies to multiple camera systems, which can share the same model through batch processing. Another factor to take into consideration is that the layers requiring FP32 data type are not supported in the DLA engine, which can only support FP16 and INT8 data types. This is not the case of FaceDetect and FaceNet which both support mixed FP16 data type.

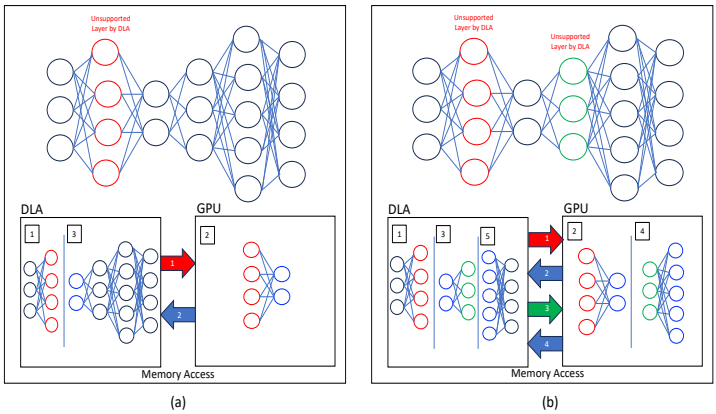

Fig. 5: Hardware Allocation with 3 partitions (a) and 5 partitions (b)

### C.3 *Minimizing the data transfers between CNN Layers*

As shown in Fig. 6, a great advantage of using DLA is that, unlike GPU, all subgraphs can be accelerated without repeated DRAM accesses. This justifies why the usage of the DLA engine leads to lower power consumption than using the CUDA cores.

Furthermore, as it will be demonstrated in Section IV, this may also lead to an increase in the system throughput as simultaneously accesses to the shared L2 memory cache and the LPDRAM5 memory are reduced. This led us to maximize the usage of the DLA engine; which is possible as long as the CNN model features a kernel in the range [1, 32] and a number of input channels in the range from 1 to 8192 [28]. This is the case for both FaceDetect and FaceNet models, which both support two-dimensional operations and a maximal kernel size of 7 x 7. Thus, another room for improvement for GPU processors is to increase the size of the memory caches (L1 and L2) to mitigate the memory conflict when leveraging pipelining.

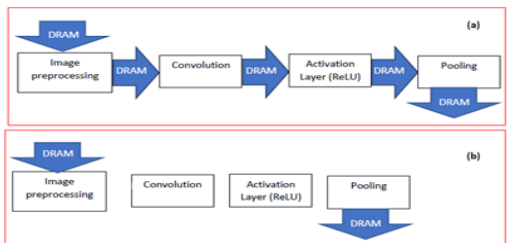

Fig. 6: Pipelining using SMs only (a) and using DLA (b)

## IV. EXPERIMENTAL RESULTS AND DISCUSSIONS

To demonstrate the merit of simultaneous utilization of all hardware engines that are available in the Orin GPUs, while satisfying the design constraints mentioned in Section III, an extensive set of experiments was done for different scenarios. An open-source video that consists of 650 frames with a total of 6 celebrity faces [8] Fig. 7, in addition to customized and lower quality videos that included up to 9 faces, were used.

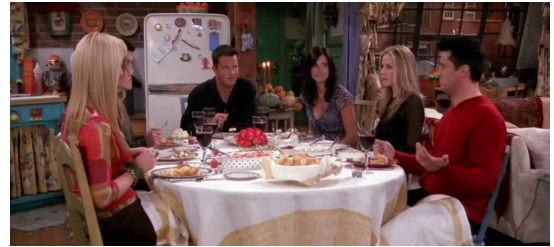

Fig. 7: Sample frame of the video used in the experiments

The purpose of the customized video is to demonstrate the system's ability to handle video streams that are not taken in ideal situations and to handle traditional headscarves from the Middle East region, which were not considered in earlier models. With regard to the open-source video, 10 faces/person were used for training to achieve a minimal accuracy of 97.8%. Similarly, 10 faces/person were used for the customized video to achieve an accuracy of 100%. Four different hardware allocation schemes that use different combinations of hardware engines were assessed for latency, throughput, and power consumption (Table I). As mentioned in Section III, the partitioning in this paper was done at the CNN model.

TABLE I
FOUR TEST RUNS ON ORIN GPU

| Run No. | Model | Target Hardware | Model | Target Hardware |
|---------|-------|-----------------|-------|-----------------|
| 1 | Face Detection (Face Detect Model) | SMs | Face Recognition (FaceNet Model) | SMs |
| 2 | | DLA 0 | | SMs |
| 3 | | SMs | | DLA 1 |
| 4 | | DLA 0 | | DLA 1 |

### A. *Latency*

Fig.1, which exhibits the frame latency for every frame, shows two distinct trends with almost a 50 ms difference. That is, running both models on SMs exhibits similar performance in terms of frame latency as running the FaceDetect on DLA and FaceNet on SMs.

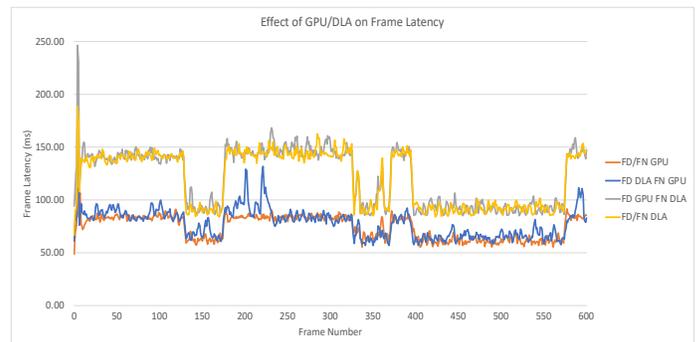

Fig. 8: Effect of Processing Unit (GPU/DLA) on Frame Latency

Similarly, running both models on the DLA gives an almost aligned frame latency trend as running the FaceDetect on GPU and FaceNet on SMs. Thus, running the FaceNet model on the DLA raised the frame latency by around 50 ms compared to running it on the GPU. This was expected because of the large number of layers (63) that are not supported by the DLA, thus forcing offloading them into the GPU and causing excessive communication between the GPU and DLA engines. These unsupported layers comprise 28 shuffle layers, 28 constant layers, global average pooling, and POW operation. These results show that the latency, which is up to 6 frames, needs to be taken into consideration in the design of the actuating part of the associated system, especially if hard real-time feedback is required.

### B. *Throughput*

Fig.9 summarizes the average throughput for the four schemes during a 2,000 ms time frame. It can be observed that all of them performed above the real-time limit of 30 FPS due to the explicit exploration of parallelism and pipelining within the processor. For instance, running the full pipeline on SMs leads to an average frame latency of around 75 ms per frame. However, the pipeline could process 192 FPS, which corresponds to 5.2 ms per frame. Similarly, the other 3 schemes could accomplish an average of 4.9 ms, 15.8 ms, and 16.1 ms per frame. It can also be observed that FaceNet model running on DLA decreases the FPS throughput of the pipeline by around 3 times compared to when it is run on the SMs. This supports the statement mentioned in Section III that many unsupported layers were introduced into the model.

Further analysis of the results indicates that with regard to the FacDetect model, 12 parameters cannot run in INT8 when enabling the DLA, and these parameters fell back to FP16 implementation. This caused some delays as the data transfer rate between the DLA and the local buffer was halved. Thus, fine-tuning the FaceDetect model to handle INT8 datatypes would have led to even higher performance.

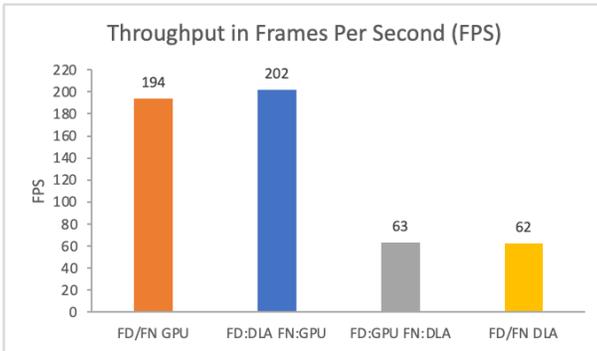

Fig. 9: Throughput performance of face detection and face recognition.

### C. *Power Consumption Performance*

To conclude which scheme leads to the best hardware allocation of the algorithm, the total power consumption also needs to be assessed. Fig. 10 shows the power consumption of both the CUDA and CPU cores, corresponding to the four schemes. It can be observed that the CUDA core power consumption when running both models on SMs is 4635 mW.

Running the FaceDetect model on DLA and the FaceNet on the CUDA core resulted in around 300 mW less power consumption, corresponding to a more than 5 % decrease. On the other hand, this hardware allocation also yields an increase of around 500 mW in the CPU core, as the CPU core is more involved in device-to-device memory transfer. This suggests that a DMA controller between the shared DRAM memory and the DLA input buffer can be useful to offload the CPU from this task. It can also be observed that schemes 3 and 4 exhibited nearly identical and highest CUDA core power consumption. This was anticipated since, as it was explained in Section III, there are some unsupported layers that fell back to the GPU.

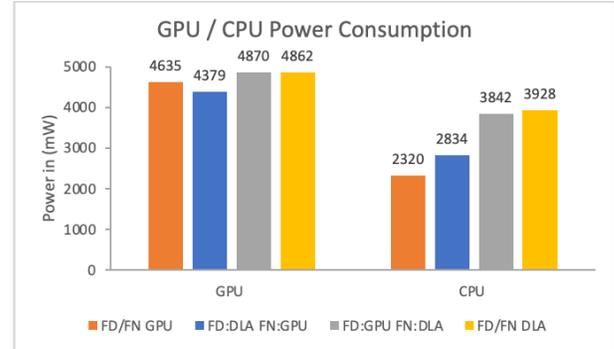

Fig. 10: CPU-GPU Power consumption of face detection and face recognition.

### D. *Effect of video decoder on the system Latency*

To assess the performance of the system in running all pipeline stages, including the video decoding stage for each of the four runs, some random video frames (e.g., frames 0, 1, 100, 200, 300, and 500) were analyzed in terms of latency (Table II). In the Table, the NVStreammux processing step was added to allow the system to accommodate multiple video streams, which, as will be shown in subsection IV-E., can enhance the system performance even further. As previously mentioned, the fastest processing time is achieved when running the face detection algorithm on the DLA and the face recognition on the GPU. In addition, it is observed that the decoder and NvStreammux stages latency increase with the increase of face recognition latency. This is not related to the performance drop of the NVDEC or CPU, but to the inferencing speed drop in the pipeline. The variable latency of the video decoder is also due to the intrinsic property of the H264 decoder as the frames are classified into I, B, and P frames. The decompression latency depends on the level of intraframe correlation (I and P, B Frames) and interframe correlations (P and B frames), which dictate how the three frames will be used to produce the decoded frame [30]. In particular, the B frame is resource-heavy and leads to the most delay, while it yields the highest compression ratio. The video decoder in the ORIN GPU generates the decoded frames out of order causing different latency for different frames.

TABLE II
PIPELINE PLUG-INS LATENCY IN (MS) FOR SELECTED FRAMES

| FD/FN GPU | | | | | | |
|---|---|---|---|---|---|---|
| **Frame No.** | **Decoder** | **NvStreammux** | **detect** | **recognize** | **Encode** | **Total** |
| **0** | 0.02 | 0.1 | 15.0 | 19.0 | 3.2 | 37.4 |
| **1** | 0.15 | 3.9 | 19.5 | 32.2 | 3.0 | 58.8 |
| **100** | 35.80 | 0.1 | 7.2 | 15.7 | 7.2 | 66.0 |
| **200** | 2.03 | 6.7 | 3.9 | 23.4 | 3.5 | 39.5 |
| **300** | 13.60 | 21.4 | 3.9 | 29.0 | 2.5 | 70.4 |
| **400** | 3.18 | 57.6 | 6.5 | 8.5 | 3.0 | 78.7 |
| **500** | 14.30 | 58.6 | 5.9 | 10.0 | 3.4 | 92.2 |
| **Average (ms)** | 9.9 | 21.2 | 8.8 | 19.7 | 3.7 | 63.3 |
| **FD_DLA FN_GPU** | | | | | | |
| **Frame No.** | **Decoder** | **NvStreammux** | **Detect** | **Recognize** | **Encode** | **Total** |
| **0** | 2.08 | 0.1 | 16.0 | 15.0 | 6.1 | 39.3 |
| **1** | 0.11 | 2.6 | 20.3 | 34.5 | 3.0 | 60.4 |
| **100** | 0.02 | 24.5 | 6.3 | 22.4 | 3.8 | 57.0 |
| **200** | 10.90 | 5.1 | 6.1 | 14.8 | 2.0 | 38.9 |
| **300** | 7.80 | 0.3 | 5.2 | 14.5 | 1.0 | 28.8 |
| **400** | 31.40 | 65.0 | 6.5 | 5.1 | 3.3 | 111.3 |
| **500** | 0.08 | 6.2 | 4.1 | 3.2 | 3.2 | 16.8 |
| **Average (ms)** | 7.5 | 14.8 | 9.2 | 15.6 | 3.2 | 50.4 |
| **FD_GPU FN_DLA** | | | | | | |
| **Frame No.** | **Decoder** | **NvStreammux** | **Detect** | **Recognize** | **Encode** | **Total** |
| **0** | 0.003 | 0.1 | 14.5 | 52.5 | 1.7 | 68.8 |
| **1** | 0.11 | 11.0 | 18.7 | 70.5 | 3.2 | 103.5 |
| **100** | 0.05 | 16.7 | 10.1 | 34.5 | 13.4 | 74.7 |
| **200** | 0.04 | 55.0 | 4.3 | 38.5 | 21.5 | 119.3 |
| **300** | 37.48 | 58.7 | 8.9 | 36.3 | 13.2 | 154.6 |
| **400** | 24.10 | 37.9 | 6.7 | 19.2 | 13.3 | 101.2 |
| **500** | 0.06 | 54.2 | 10.7 | 19.7 | 13.4 | 98.1 |
| **Average (ms)** | 8.8 | 33.4 | 10.6 | 38.7 | 11.4 | 102.9 |
| **FD/FN DLA** | | | | | | |
| **Frame No.** | **Decoder** | **NvStreammux** | **Detect** | **Recognize** | **Encode** | **Total** |
| **0** | 2.03 | 0.2 | 15.9 | 25.3 | 4.1 | 47.6 |
| **1** | 0.01 | 10.6 | 18.9 | 48.3 | 18.8 | 96.6 |
| **100** | 31.30 | 15.9 | 17.3 | 66.5 | 21.2 | 152.2 |
| **200** | 0.04 | 54.8 | 5.8 | 30.5 | 21.4 | 112.5 |
| **300** | 19.70 | 55.1 | 4.2 | 41.3 | 13.5 | 133.8 |
| **400** | 24.70 | 39.0 | 4.1 | 18.6 | 20.1 | 106.5 |
| **500** | 20.50 | 61.4 | 8.9 | 17.4 | 13.2 | 121.4 |
| **Average (ms)** | 14.0 | 33.9 | 10.7 | 35.4 | 16.0 | 110.1 |

### E. *Leveraging batch processing*

To evaluate the performance of the pipeline for multiple video sources, two video streams were simultaneously introduced into the pipeline. The Gst-Nvstreammux module, which is run on the CPU, controls the flow of the two video streams between the LPDRAM5 memory and the pipeline hardware. Its round-robin scheduling policy allowed the two video streams to be processed approximately at the same throughput Fig. 11 (a). It can be observed that both single and multi-streams running the face detection on DLA result in a better throughput than running the same model on the GPU.

Fig. 11(b) shows that processing a single video source requires slightly more time than processing double video sources due to the higher GPU utilization. It could also be observed that doubling the number of video sources does not increase the CPU/GPU power consumption Fig. 11(c).

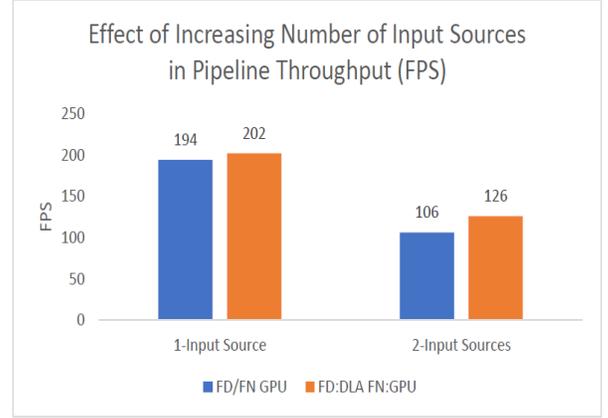

(a)

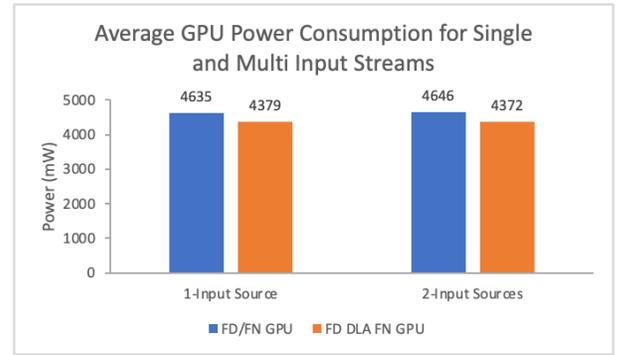

(b)

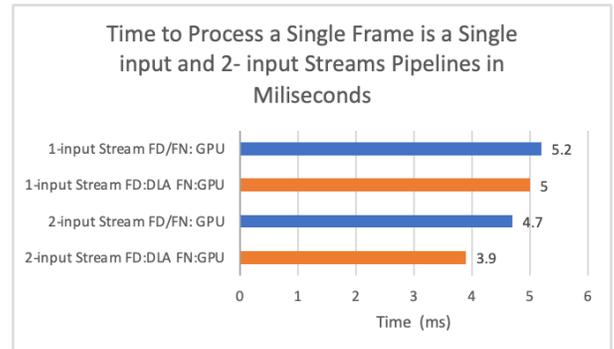

(c)

Fig. 11: Effect of Increasing Number of Input Sources in Pipeline Throughput (a) latency (b) and power (c)

This can be explained by the fact that doubling the number of video sources raised the GPU load by only 8% compared to a single video source. However, even with this increase, the GPU utilization load is still around 40%; which still falls within the GPU capacity, and therefore, it does not result in a significant power consumption increase. The same trend was observed with regard to the CPU power consumption

### G. *Discussions*

Among the most recent similar works, one can list the work presented in [9], where the MTCNN model, which yields the same accuracy as the FaceDetect model of at least 97 %, was used for face detection. Similarly to our work, Facenet was used for face recognition. The two algorithms, which were implemented on edge GPU (e.g., Jetson AGX) and cloud GPU (e.g., RTX 2080 ti), were tested on a video sequence, and it was not mentioned if the video decoding task was implemented within the dedicated ASIC module of the GPU. They used three different sizes of the input videos (i.e., 480x480, 1280 x 720, and 1920 x 1080 pixels, respectively), whereas in our case, only 1920 x 1080 input video streams were considered. Also, on average, there were around 3.4 faces/image, whereas in our case, there were around 4 faces on average to have up to 9 faces/frame. The authors solely used the GPU CUDA cores using the TensorRT framework (in addition to the lighter TFLite and the unoptimized TEnsorFLow models) without leveraging the usage of the other hardware engines, as suggested in this paper. Our implementation yielded a shorter delay than their Jetson AGX implementation for all four schemes, where an average latency of 180 ms was achieved using the TensorRT framework. On the other hand, their implementation of the two models on the RTX2080ti GPU yielded, on average, a shorter latency of 30 ms; however, at the cost of an excessive power consumption of up to 200 W. This was significantly higher than our implementation, where the maximal GPU power consumption was less than 5 W in all four schemes, slightly higher than their edge GPU implementation, where it reached up to 2.3 W. Nevertheless, the throughput of their system on both the edge and cloud GPU, which was 1.75 FPS and 7.79 FPS respectively, was much lower. This demonstrates the merit of exploring all the hardware engines of the GPU and promoting the pipelining architecture.

## V. CONCLUSION

This paper demonstrates that the simultaneous usage of different hardware engines that are available within modern edge GPUs can enhance the system's performance in terms of throughput and latency while consuming less power in the CUDA core module. However, the total power consumption gets slightly higher when considering the CPU engine, as further CPU workload is required to manage the DLA memory hardware engine. It also demonstrates that increasing the batch size to two video streams leads to higher performance with approximately the same power consumption as processing a single video stream. The performance could have been even slightly better if each video stream was handled by separate DLA hardware engines, which is not possible to do in the current NVIDIA framework. Four different hardware allocations support this claim by assessing the corresponding latency, throughput, and power consumption for single and multiple video streams. The best performance was achieved by running the face detection model on the DLA engine and the face recognition model on the CUDA core, obtaining a throughput of 202 FPS. This constitutes an improvement over other previous hardware accelerators, which implemented the face detection and recognition algorithms separately. This achievement can potentially promote the development of other similar emerging applications, such as real-time machine vision for autonomous vehicles and big data processing acceleration. A reasonable number of faces in the image ranging from 3 to 9 faces and taken from a public dataset, as well as customized videos taken in the lab, were considered in the experiments. This is better than other related works, which considered only up to 3 faces. In addition, some of the videos that were used in the assessment include traditional headscarves from the Middle East region, which were not considered in earlier models. Following the experimental results, the paper suggests some hardware improvements of the existing edge GPU processors to enhance their performance even better. For instance, adding a shuffle layer accelerator within the DLA engine would avoid offloading this task onto the GPU, which highly degraded the performance of the face recognition model onto the DLA. This allows seamless implementation of existing CNN models into DLA, using high-level frameworks, without compromising the system's performance. Another room for improvement in current NVIDIA frameworks is to provide an automated hardware-software co-design scheme at the layer level using, for instance, Equations (1) and (2) above to ensure an optimal hardware allocation. For instance, In the case of the face detection/recognition algorithm, L2 memory cache capacity is enough to store two video frames, which are processed by two SM pipeline stages (e.g. face detection and face recognition pipeline stages) of the same GPU. This lowers the memory conflict between the two SMs, which enhances the throughput performance compared to the case when using DLA for implementing the face recognition stage. Thus, another room for improvement for GPU processors is to increase the size of the memory caches (L1 and L2) to mitigate the memory conflict when leveraging pipelining. Another not less significant room for improvement is to offload the multicore CPU from the memory transfer tasks between the DLA buffer and the shared memories using direct memory access controllers. This would allow the multicore CPU to be allocated for other more computation-intensive tasks.